\documentclass{ecai}

\usepackage{latexsym}
\usepackage{amssymb}
\usepackage{amsmath}
\usepackage{amsthm}
\usepackage{booktabs}
\usepackage{enumitem}
\usepackage{graphicx}
\usepackage{color}
\usepackage{colortbl}
\usepackage[most]{tcolorbox}
\usepackage{multirow}
\usepackage{siunitx}

\newcommand{\BibTeX}{B\kern-.05em{\sc i\kern-.025em b}\kern-.08em\TeX}
\newcommand{\code}[1]{{\footnotesize\ttfamily#1}}

\begin{document}

\begin{frontmatter}

\paperid{8}

\title{Optimizing Small Transformer-Based Language Models for Multi-Label Sentiment Analysis in Short Texts}

\author[A]{\fnms{Julius}~\snm{Neumann}}
\author[A]{\fnms{Robert}~\snm{Lange}}
\author[B]{\fnms{Yuni}~\snm{Susanti}} 
\author[A]{\fnms{Michael}~\snm{Färber}\thanks{Corresponding Author. Email: michael.faerber@tu-dresden.de}}

\address[A]{ScaDS.AI, TU Dresden, Germany}
\address[B]{FIZ Karlsruhe, Germany}

\begin{abstract}
Sentiment classification in short text datasets faces significant challenges such as class imbalance, limited training samples, and the inherent subjectivity of sentiment labels—issues that are further intensified by the limited context in short texts. These factors make it difficult to resolve ambiguity and exacerbate data sparsity, hindering effective learning.
In this paper, we evaluate the effectiveness of \textit{small} Transformer-based models (i.e., BERT and RoBERTa, with fewer than 1 billion parameters) for multi-label sentiment classification, with a particular focus on short-text settings. Specifically, we evaluated three key factors influencing model performance: (1) continued domain-specific pre-training, (2) data augmentation using automatically generated examples, specifically \textit{generative data augmentation}, and (3) architectural variations of the classification head. Our experiment results show that data augmentation improves classification performance,
while continued pre-training on augmented datasets can introduce noise rather than boost accuracy. Furthermore, we confirm that modifications to the classification head yield only marginal benefits. %
These findings provide practical guidance for optimizing BERT-based models in resource-constrained settings and refining strategies for sentiment classification in short-text datasets. 
\end{abstract}

\end{frontmatter}

\section{Introduction}
In modern practical decision-making, building predictive models often involves characterized by significant challenges, including imbalanced class distributions, highly variable dataset sizes—from limited to extremely large—and inherent subjectivity in the ground truth labels. These factors can severely degrade model performance and further complicate learning. Consequently, developing novel learning methods capable of addressing these difficulties is a critical area of ongoing research. Within this context, multi-label sentiment classification in short texts represents a particularly challenging task, requiring models to infer subjective emotional states (e.g., anger, joy, or surprise) from sparse and ambiguous linguistic cues. Short texts inherently limit contextual information, reducing the availability of disambiguating semantic cues and amplifying the effects of data sparsity. Additional common issue in this task includes inherently subjective nature of sentiment annotations, which often results in low inter-annotator agreement.
Furthermore, short texts—especially from social media—frequently contain informal language, abbreviations, emojis, and sarcasm, further complicating accurate sentiment detection. Addressing these compounded difficulties is essential for advancing sentiment analysis and related Natural Language Processing (NLP) tasks and applications.

Although pre-trained Transformer-based models such as BERT~\cite{devlin2019bertpretrainingdeepbidirectional} have significantly advanced sentiment analysis \cite{Alaparthi2020BidirectionalER, devlin2019bertpretrainingdeepbidirectional, liu2019robertarobustlyoptimizedbert}, critical challenges remain, including handling class imbalance, data scarcity, and subtle semantic cues within short texts. Recent research has shown that expanding training data through generative data augmentation \cite{Balkus_Yan_2024}, refining model architectures \cite{stickland2019bertpalsprojectedattention}, and using domain-specific pre-training \cite{Ji2023ExploringTI} potentially enhance the model's performance. However, only few studies systematically evaluate these approaches in combination. Existing research predominantly focuses on binary sentiment classification or assumes access to large, well-curated datasets, leaving open questions about the optimal strategies for multi-label emotion detection in small, noisy and inherently subjective text corpora.

In this paper, we address these challenges through a comprehensive evaluation of the following three key strategies: (1) continued domain-specific pre-training to better capture linguistic nuances, (2) generative data augmentation (GDA) to alleviate data scarcity, and (3) classification head variations to enhance parameter efficiency and adaptability. Our experiments were carried out on 2,768 short English texts from the SemEval 2025 Task 11 corpus~\cite{codabench_challenge}, utilizing \textit{small} Transformer-based language models, specifically BERT~\cite{devlin2019bertpretrainingdeepbidirectional} and its variant RoBERTa~\cite{liu2019robertarobustlyoptimizedbert}. In this work, we define \textit{small} as language models with fewer than 1 billion parameters, following~\cite{kgprompt-10.1007/978-3-031-77844-5_5}. Such models offer a favorable trade-off between performance and computational efficiency, making them well-suited for deployment in resource-constrained real-world scenarios. 
Additionally, we apply SHAP~\cite{NIPS2017_8a20a862} analysis to improve the interpretability of the predictions. 

The following summarizes our main contributions:
\begin{itemize}
    \item We systematically evaluate three strategies—continued domain-specific pre-training, generative data augmentation, and classification head variations—for multi-label sentiment classification in short texts, targeting challenges of linguistic sparsity, data scarcity, and model efficiency. We release the code and dataset on Github.\footnote{https://github.com/faerber-lab/shorttext-sentiment-transformers}
    \item Our results show that moderate generative data augmentation improves BERT-based model performance, while excessive domain-specific pre-training can add noise. Variations in classification heads yield minimal gains. These findings provide practical guidance for BERT optimization in real-world, short-text datasets.
    \item We employ SHAP analysis to interpret model predictions, offering insights into feature contributions and enhancing model transparency in short-text sentiment classification tasks.
\end{itemize}

\section{Related Work}
\paragraph{Transformer-Based Text Classification.}
Transformer-based models such as BERT~\cite{devlin2019bertpretrainingdeepbidirectional} have become the state of the art for text classification, largely due to their ability to encode contextualized representations with bidirectional attention \cite{Alaparthi2020BidirectionalER}. Compared to earlier architectures like LSTMs, Transformers more effectively capture long-range semantic dependencies. Beyond model architecture, the quantity of available training data strongly influences performance. \citeauthor{Ji2023ExploringTI} \cite{Ji2023ExploringTI} and \citeauthor{Mehrafarin2022OnTI} \cite{Mehrafarin2022OnTI} show that larger training sets generally lead to improved generalization in NLP tasks. BERT \cite{devlin2019bertpretrainingdeepbidirectional} and its refined variant RoBERTa \cite{liu2019robertarobustlyoptimizedbert} are often the starting points for a wide range of downstream applications. RoBERTa’s removal of next sentence prediction and its employment of extended pre-training with dynamic masking enhance performance on multiple benchmarks \cite{liu2019robertarobustlyoptimizedbert, Hirlea2021ContextualSC}. In typical classification scenarios, the hidden representation of the \code{[CLS]} token is passed through a classifier, and both the Transformer layers and the classifier head are fine-tuned to maximize the probability of the correct label. Adjustments in the head’s architecture, such as varying embedding size or adopting parameter-efficient layers like \emph{Projected Attention} \cite{stickland2019bertpalsprojectedattention}, can further enhance performance \cite{Wolfe2021ExceedingTL}.

\paragraph{Data Augmentation.}
Data augmentation has been widely adopted to improve model generalization in NLP tasks. It typically involves selecting or creating meaningful samples, such as work by~\cite{susanti2024dataaugmentationtechniquesprocess} or adding extra information \cite{guo2022domainadaptationgeneralizationpretrained}. While adversarial approaches have been explored \cite{guo2022domainadaptationgeneralizationpretrained, qu2020codacontrastenhanceddiversitypromotingdata}, their complexity and mixed empirical results have prompted interest in simpler, generative methods. \citeauthor{Balkus_Yan_2024} \cite{Balkus_Yan_2024} propose a more straightforward strategy, using GPT-3 to generate new training examples and filtering them to retain only the most promising candidates. Even with a dataset of only 26 examples and two labels, they report substantial accuracy improvements, but also observe that excessive augmentation can introduce noise and ultimately degrade performance \cite{10.1145/3486622.3493960}.

\paragraph{Generative Data Augmentation.}
Although generative data augmentation (GDA) has been explored in previous work \cite{ISWC2024workshop}, we adapt it here for multi-label sentiment classification on short texts. In contrast to Unsupervised Data Augmentation (UDA) \cite{xie2020unsuperviseddataaugmentationconsistency}, which emphasizes consistency training on unlabeled data, we directly generate new examples using a state-of-the-art language model (e.g., GPT-4o) and incorporate them into the training set without additional filtering. We refer to this approach as \textit{Generative Data Augmentation (GDA)}. 
By verifying that the synthetic samples remain domain-consistent, we aim to capture rare sentiment cues and edge cases often missing from the original dataset. Similar to prior studies, we hypothesize that augmenting too aggressively may introduce noise and lead to performance plateaus or overfitting, but a moderate infusion of diverse synthetic data can substantively boost our multi-label classification performance.

\section{Methodology}
Our work focuses on investigating the performance of Transformer-based \textit{small} language models for short-text multi-label sentiment analysis task. Given a short text as the input, the model predicts one or more sentiment labels that best describe the emotional content of the input text. Formally, let $\mathcal{D} = \{(x_i, Y_i)\}_{i=1}^{N}$ represent the dataset, where each sample consists of: 
\begin{itemize}[leftmargin=0.7cm]
    \item $x_i \in \mathcal{X}$: a short text sample as the input
    \item $Y_i \subseteq \mathcal{Y}$: a subset of sentiment labels assigned to $x_i$, where $\mathcal{Y} = \{\text{Anger}, \text{Fear}, \text{Joy}, \text{Sadness}, \text{Surprise}\}$ is the set of all possible sentiment classes.
\end{itemize}

Unlike single-label classification, each instance in the dataset can be associated with multiple sentiment labels simultaneously. To evaluate model performance, we employ standard multi-label classification metrics following \citeauthor{codabench_challenge}~\cite{codabench_challenge}, including \textit{Accuracy}, \textit{F1 Score}, \textit{Macro F1 Score}, and \textit{Cohen’s Kappa Score}. Macro F1 Score metric ensures a balanced assessment across all sentiment categories, addressing potential class imbalance issues.

\subsection{Approach}
In this work, we investigate three distinct approaches to enhance the performance of Transformer-based language models,
described in details below. Our experiments focus on BERT and its variant, RoBERTa, as the base models.

\subsubsection{Continued Domain-Specific Pre-training}
This approach involves continuing the pre-training of base language models on domain-specific corpora, as recommended by \citeauthor{sun2020finetuneberttextclassification} \cite{sun2020finetuneberttextclassification}. This strategy is particularly relevant for short-text sentiment classification, where limited context and sparse linguistic cues make capturing domain-specific language nuances critical. Supporting this, \citeauthor{10.1145/3486622.3493960} \cite{10.1145/3486622.3493960} demonstrated that domain-focused pre-training leads to significantly greater performance gains than merely increasing the size of the general training dataset beyond a certain threshold, highlighting the importance of specialized knowledge for improving accuracy in contextually constrained scenarios such as short texts.

To perform continued pre-training of the models in our experiment, only the \textit{text} field of the training dataset is utilized, following the work by~\citeauthor{sun2020finetuneberttextclassification}~\cite{sun2020finetuneberttextclassification}. The text samples are tokenized using a pre-trained tokenizer, with each model employing its corresponding tokenizer from \code{HuggingFace}~\cite{hugDBLP:journals/corr/abs-1910-03771} library. The models are then trained on these tokenized text samples using the masked language modeling (MLM) objective, wherein tokens are randomly masked with a probability of 15\%. Pre-training is carried out for 50 epochs with a learning rate of $2 \times 10^{-5}$ and a weight decay of 0.01.

\subsubsection{Generative Data Augmentation (GDA)} 
This approach enhances the original dataset by incorporating automatically generated samples from a generative model, which is particularly beneficial for our task where data scarcity pose significant challenges. By expanding the training data with synthetic examples that reflect the original dataset-specific language patterns, GDA helps models better capture subtle linguistic nuances that are often underrepresented in particularly small datasets. Prior studies have demonstrated that such augmentation can significantly improve model performance in certain cases \cite{Ji2023ExploringTI, Mehrafarin2022OnTI}. 

To augment the original training set in our experiment, we utilized OpenAI completion endpoint with a fine-tuned \code{GPT-4o-mini model}. This model was fine-tuned for three epochs on a partitioned training set incorporating sentiment labels, to match the label distribution of the dataset under test. The system prompt specified the following instruction:

\begin{tcolorbox}
    [
    colback=gray!5, 
    colframe=black, 
    fonttitle=\bfseries,
    boxsep=3pt, 
    left=3pt, 
    right=3pt, 
    top=3pt, 
    bottom=3pt,
    sharp corners ]
Generate short texts with their corresponding sentiment labels. The sentiment labels include Anger, Fear, Joy, Sadness, and Surprise. The texts are in English and have a maximum length of 256 characters.  
\end{tcolorbox}

For each generated example, we provided the same prompt along with an output template in a \code{JSON} schema, which included a text field and an array of assigned sentiment labels. This procedure was repeated until a total of 11,684 augmented examples were generated, a number chosen to align with our available computational resources. Since the generated texts exhibited slight variations in average length compared to the original dataset, many samples were split into two or three parts to better align with the original training domain. Subsequent fine-tuning on the augmented dataset resulted in improved model performance, reducing label ambiguity, and confirming that the generated data maintained a quality comparable to that of the original dataset. The class distribution of the original dataset and the dataset after augmentation using the generative data augmentation method are presented in Tables~\ref{tab:label_stats} and ~\ref{tab:label_stats_GDA}, respectively.

\subsubsection{Classification Head Architectures}
In the third approach, we evaluate different architectures for the model’s classification head, a component especially critical for short-text sentiment classification where limited input length demands efficient extraction and integration of the model features. As highlighted by \citeauthor{Verma2021AttentionIA}~\cite{Verma2021AttentionIA}, the design of the feed-forward classification head can substantially impact overall model performance. We compare two distinct classification head architectures: (1) a \textit{fully-connected layer} and (2) \textit{projected attention}, described in detail below.

\begin{figure}[tb]
    \centering
    \includegraphics[width=0.65\linewidth]{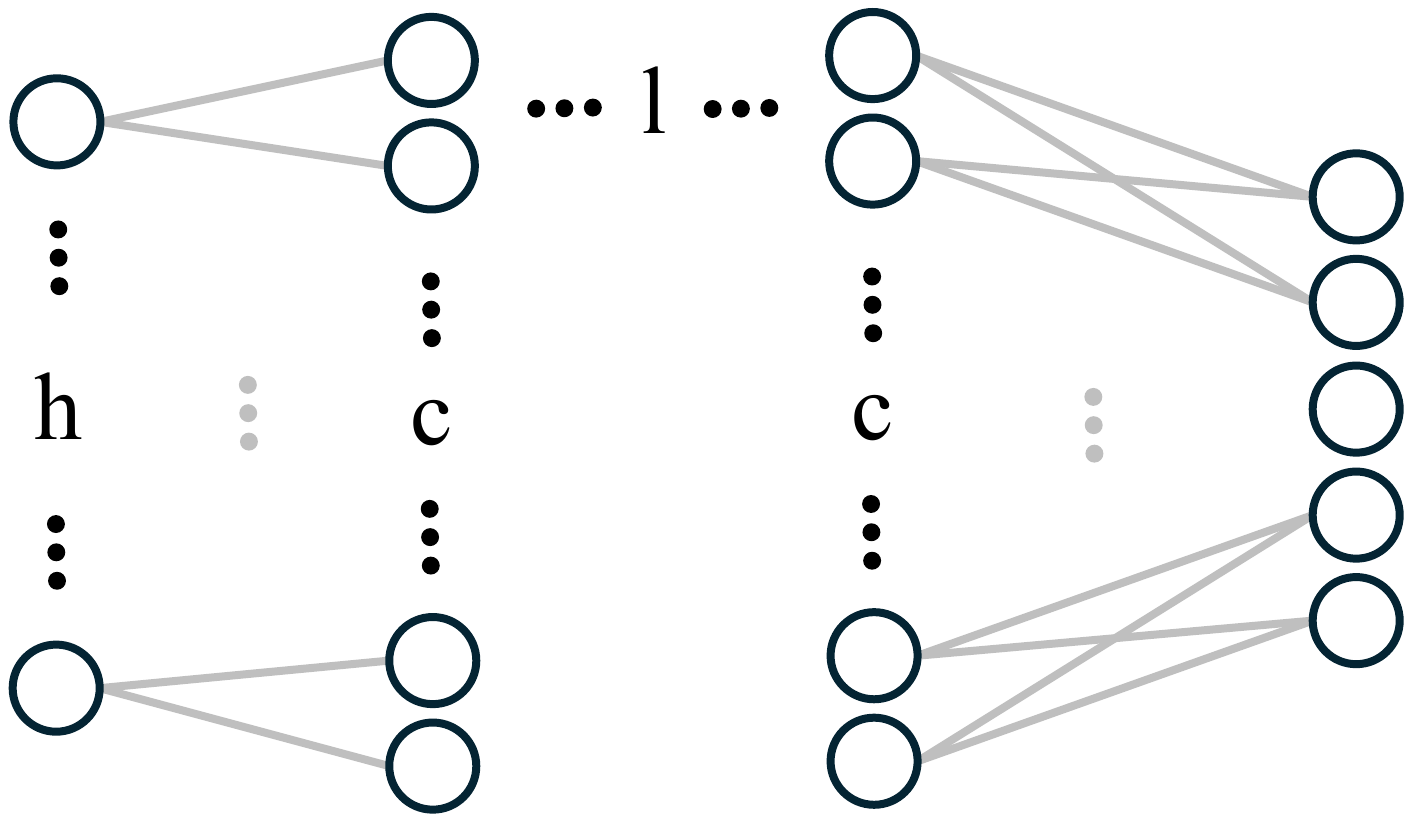}
    \caption{Illustration of the \textit{fully-connected layer} classification head.}
    \label{fig:Fully Connected Head}
\end{figure}

A \textbf{\textit{fully-connected layer}} architecture provides a straightforward mapping from encoded features to labels. Concretely, it consists of a fully-connected layer network with intermediate \code{ReLU} activation function and dropout layers, as illustrated in Figure~\ref{fig:Fully Connected Head}. In this setting, the hidden size $h$ of the pre-trained model serves as the input dimension. This is followed by $l$ linear layers, each with an internal classifier size $c$. The final classification layer produces an output of size equal to the number of labels. Finally, a \textit{sigmoid} function is applied to compute the probability for each label, following the practice in multi-label classification tasks. 

\begin{figure}[tb]
    \centering
    \includegraphics[width=0.65\linewidth]{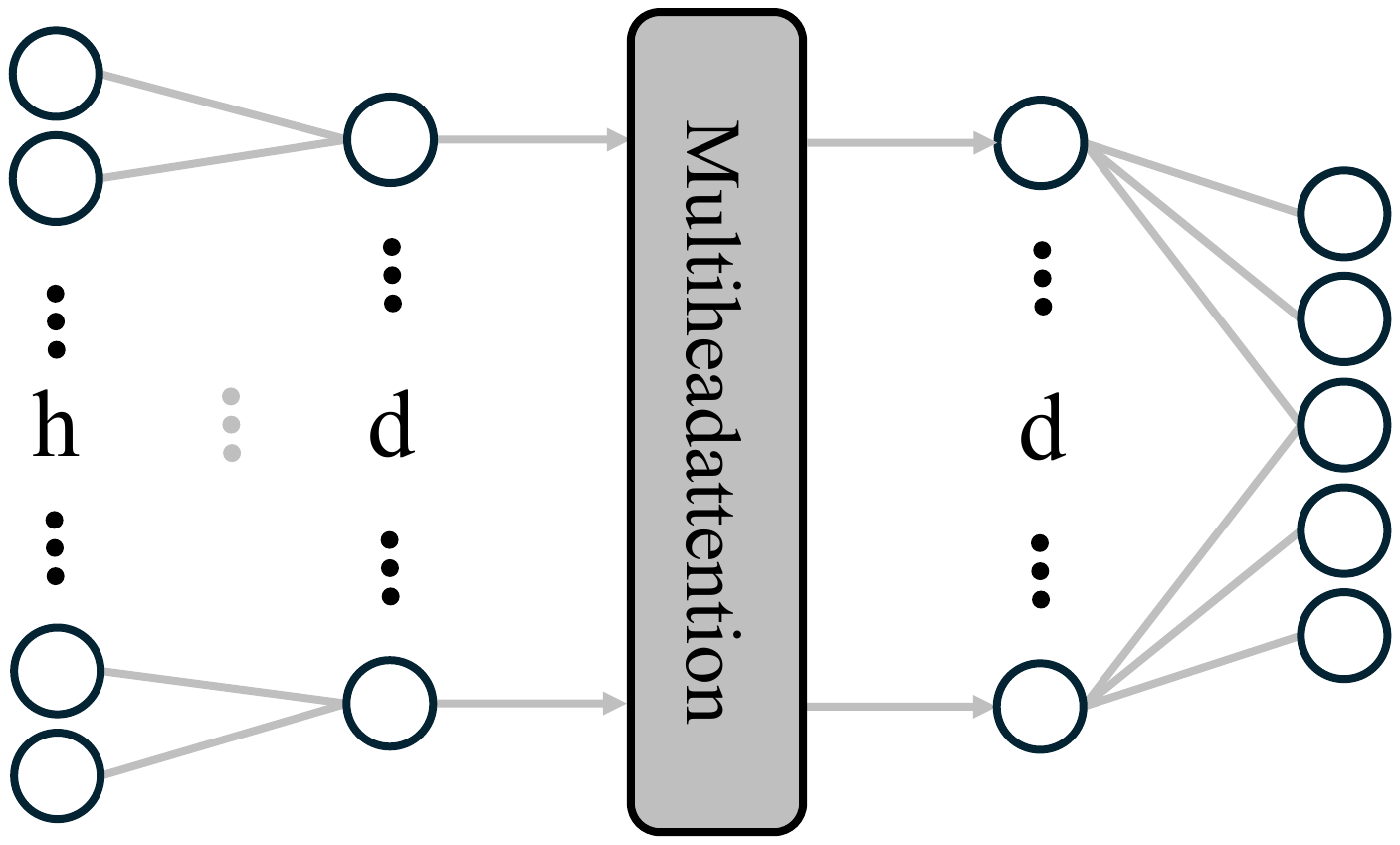}
    \caption{Illustration of the \textit{projected attention} classification head.}
    \label{fig:Attention Projection}
\end{figure}

The second architecture employs the \textbf{\textit{projected attention}} mechanism, as proposed by \citeauthor{stickland2019bertpalsprojectedattention}~\cite{stickland2019bertpalsprojectedattention}. This technique incorporates an attention mechanism (see Figure~\ref{fig:Attention Projection}) to dynamically focus on important features—potentially enhancing the model’s ability to capture nuanced sentiment signals in sparse, context-limited short texts. In this architecture, the classification head consists of a linear projection layer that transforms the pre-trained model's output size $h$ to the attention dimension $d$. This is followed by a multi-head attention mechanism with $k$ attention heads and an embedding dimension of $d$. Lastly, an attention-based classifier projects the attention output to match the number of labels.

\begin{table}[tb]
\centering
\begin{tabular}{l S[table-format=4.0] S[table-format=1.2]}
\toprule
\textbf{Label} & \textbf{Frequency} & \textbf{Probability (\%)} \\
\midrule
Anger      & 333   & 12.0 \\
Fear       & 1611  & 58.2 \\
Joy        & 674   & 24.3 \\
Sadness    & 878   & 31.7 \\
Surprise   & 839   & 30.3 \\
\bottomrule
\end{tabular}
\caption{Original dataset class distribution.}
\label{tab:label_stats}
\end{table}

\begin{table}[tb]
\centering
\begin{tabular}{l S[table-format=4.0] S[table-format=1.2]}
\toprule
\textbf{Label} & \textbf{Frequency} & \textbf{Probability (\%)} \\
\midrule
Anger      & 2227   & 19.0 \\
Fear       & 7606  & 65.0 \\
Joy        & 3577   & 30.6 \\
Sadness    & 4884   & 41.8 \\
Surprise   & 4645   & 39.7 \\
\bottomrule
\end{tabular}
\caption{Augmented dataset class distribution.}
\label{tab:label_stats_GDA}
\end{table}

\subsection{Dataset}
Our experiments are conducted on the SemEval 2025 Task 11 dataset~\cite{codabench_challenge}, comprises 2,768 examples where each containing short English texts labeled with zero, one, or multiple sentiment classes: \{Anger, Fear, Joy, Sadness, and Surprise\}. On average, each text consists of 78.4 characters, corresponding to approximately 15 words, which categorizes this task as a ``short text'' classification task.

The distribution of labeled classes in the training set is presented in Table~\ref{tab:label_stats}. The dataset exhibits a class imbalance, with the ``Fear'' class being over-represented (58.2\% of the total instances) and the ``Anger'' class being under-represented (12.0\% of the total instances). This may introduce bias in the classification model, potentially leading to a higher propensity for predicting the over-represented class (``Fear'' class) and a reduced likelihood of correctly identifying instances of the under-represented class (``Anger'' class). The class distribution of the dataset after augmented with generative data augmentation method is presented in Table~\ref{tab:label_stats_GDA}. 

\subsection{Experimental Details}
In the experiment, we consider four different variants of BERT and RoBERTa pre-trained models as base models for the task-specific fine-tuning: (1) \code{bert-base}, (2) \code{bert-large}, (3) \code{roberta-base}, and (4) \code{roberta-large}. Each model is initialized from a checkpoint available in \code{HuggingFace}~\cite{hugDBLP:journals/corr/abs-1910-03771}. This setup enables a comparative analysis of BERT and RoBERTa model architectures while also assessing the impact of model size.  

In each training configuration, all four models are fine-tuned twice: once without continued pre-training and once with additional pre-training on domain-specific data. In the first configuration, the models are pre-trained and fine-tuned on the original training dataset provided by \citeauthor{codabench_challenge}~\cite{codabench_challenge}. In the second configuration, 33.3\% of the artificially generated samples are incorporated to expand the training dataset. This process is then repeated twice, once with 66.6\% of the generated samples and again with 100\% of the generated samples, to analyze whether performance improves further or diminishes. Before tokenization, all words in the training dataset are converted to lowercase, and hyperlinks as well as punctuation marks are removed.  

For all configurations, the classification head consists of a fully connected network with $l = 1$ linear layer and an internal classifier size of $c = 768$, which matches the hidden size of the BERT models. To determine whether a different classification head size or architecture improves model performance, the best-performing base model from the initial trials is further fine-tuned with a fully connected classification head containing $l = 2$ and $l = 4$ linear layers. Additionally, fine-tuning is performed with a "Projected Attention" head incorporating $k = 1$ and $k = 2$ attention heads.  

All models are fine-tuned for 30 epochs with a learning rate of $2 \times 10^{-5}$, a weight decay of 0.01, \code{AdamW} optimizer, and a training and evaluation batch size of 32. All experiments were conducted on NVIDIA A100 and V100 GPUs, using the single, predefined train-validation-test split specified by the original dataset to ensure consistency with prior works. Other hyperparameter setting and implementation details are provided in our Github.

\subsection{Explainability and Reasoning Metrics}
To further understand the reasoning behind model predictions, we generate confusion matrices and employ explainability library \code{SHAP} (SHapley Additive exPlanations)~\cite{NIPS2017_8a20a862} to analyze feature importance in classification decisions. Confusion matrix analysis helps identify common misclassifications and potential biases, providing valuable information about the limitations of the models' decision-making process. Meanwhile, SHAP utilizes \textit{Shapley} values to compute feature contribution scores, enabling a detailed assessment of input feature importance. By applying SHAP analysis, we can determine the impact of specific words on sentiment classification, offering a more interpretable understanding of model behavior.

\section{Evaluation}

We conducted evaluation experiments to assess the impact of the three strategies outlined in Section 3.1, \textit{Approach}. Additionally, we performed an explainability analysis to gain deeper insights into the model’s decision-making process. The main experimental findings and explainability results are presented in the following sections.

\begin{table*}[tb]
    \centering
    \footnotesize
    \begin{tabular}{ll|cc|cc|cc|cc}
        \toprule
        \multirow{2}{*}{\textbf{Model}} & \multirow{2}{*}{\textbf{Pretrained}} 
        & \multicolumn{2}{c|}{\textbf{Original Dataset}} 
          & \multicolumn{2}{c|}{\textbf{0.33 Augmentation}} 
          & \multicolumn{2}{c|}{\textbf{0.66 Augmentation}} 
          & \multicolumn{2}{c}{\textbf{1.0 Augmentation}} 
          \\ \cmidrule(lr){3-4} \cmidrule(lr){5-6} \cmidrule(lr){7-8} \cmidrule(lr){9- 10}
         & 
          & \textbf{Acc.} & \textbf{F1 Score} 
          & \textbf{Acc.} & \textbf{F1 Score} 
          & \textbf{Acc.} & \textbf{F1 Score} 
          & \textbf{Acc.} & \textbf{F1 Score} \\
        \midrule
        \multirow{2}{*}{\code{bert-base}} 
          & Yes & 0.365 & 0.694 & 0.375 & 0.687 & 0.365 & 0.71 & 0.438 & 0.706 \\
          & No  & 0.344 & 0.677 & 0.313 & 0.686 & 0.406 & 0.715 & 0.354 & 0.666 \\
        \midrule
        \multirow{2}{*}{\code{bert-large}} 
          & Yes & 0.427 & 0.697 & 0.406 & 0.708 & 0.375 & 0.697 & - & - \\
          & No  & 0.354 & 0.689 & 0.365 & 0.698 & 0.385 & 0.703 & 0.406 & 0.715 \\
        \midrule
        \multirow{2}{*}{\code{roberta-base}} 
          & Yes & 0.417 & 0.707 & 0.365 & 0.678 & 0.406 & 0.693 & 0.427 & 0.719 \\
          & No  & 0.396 & 0.711 & 0.396 & 0.698 & 0.406 & 0.706 & 0.458 & 0.749 \\
        \midrule
        \multirow{2}{*}{\code{roberta-large} }
        & Yes & 0.448 & 0.73 & 0.406 & 0.74 & 0.417 & 0.736 & 0.396 & 0.733 \\
        & No  & \textbf{0.469} & \textbf{0.77} & 0.396 & 0.719 & \textbf{0.469} & 0.764 & 0.458 & 0.767 \\
        \bottomrule
    \end{tabular}
    \caption{Model performance across different continued pre-training variations and data augmentation.}
    \label{tab:model_performance}
\end{table*}

\subsection{Main Results}

\begin{figure}[tb]
  \centering
  \makebox[\linewidth]{\hspace{-0.05\linewidth}\includegraphics[angle=-90,width=1.0\linewidth]{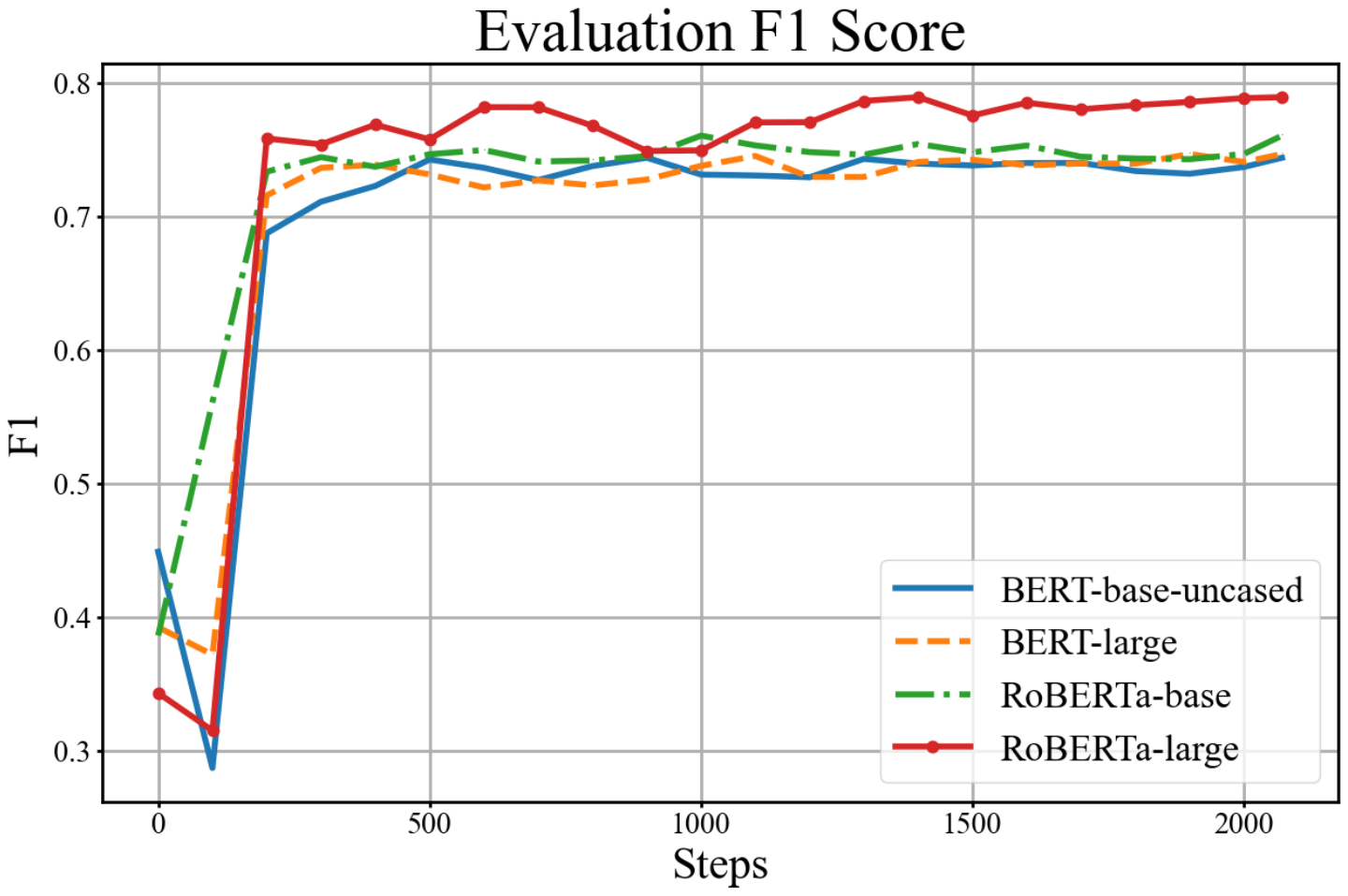}}
  \caption{F1 Scores of the four base models with continued pre-training and fine-tuning on the original dataset.}
  \label{fig:Evaluation logs}
\end{figure}

The evaluation results for continued pre-training of the models and data augmentation techniques are summarized in Table~\ref{tab:model_performance} and illustrated in Figure \ref{fig:Evaluation logs}, while the experiment results for classification head architecture variation are presented in Table~\ref{tab:roberta_large_heads_normal_vertical}. 

As shown in Figure \ref{fig:Evaluation logs}, after approximately 500 training steps, the F1 Score of the \code{bert-base}, \code{bert-large}, and \code{roberta-base} models plateaus, indicating minimal further improvement. In contrast, the \code{roberta-large} model continues to benefit from extended fine-tuning. Additionally, while larger models generally achieve better performance, the performance gap between \code{bert-base} and \code{bert-large} on the unmodified training set is marginal, with \code{bert-base} even outperforming \code{bert-large} in one instance. Conversely, the \code{roberta-large} model consistently outperforms \code{roberta-base} and is generally the best-performing model overall.

Overall, although larger models, such as \code{bert-large} and \code{roberta-large}, generally perform better, smaller models can be improved effectively through data augmentation and domain-specific pre-training. Among these, data augmentation proved most effective, improving all models except \code{roberta-large}. Modifying the classification head had little impact on performance but may reduce trainable parameters, which is beneficial in resource-constrained settings. In the following, we discuss the impact of each of the three approaches more in detail.

\paragraph{Impact of the Continued Domain-Specific Pre-Training.}

 \begin{figure}[tb]
  \centering
  \makebox[\linewidth]{\hspace{-0.05\linewidth}\includegraphics[angle=-90,width=1.0\linewidth]{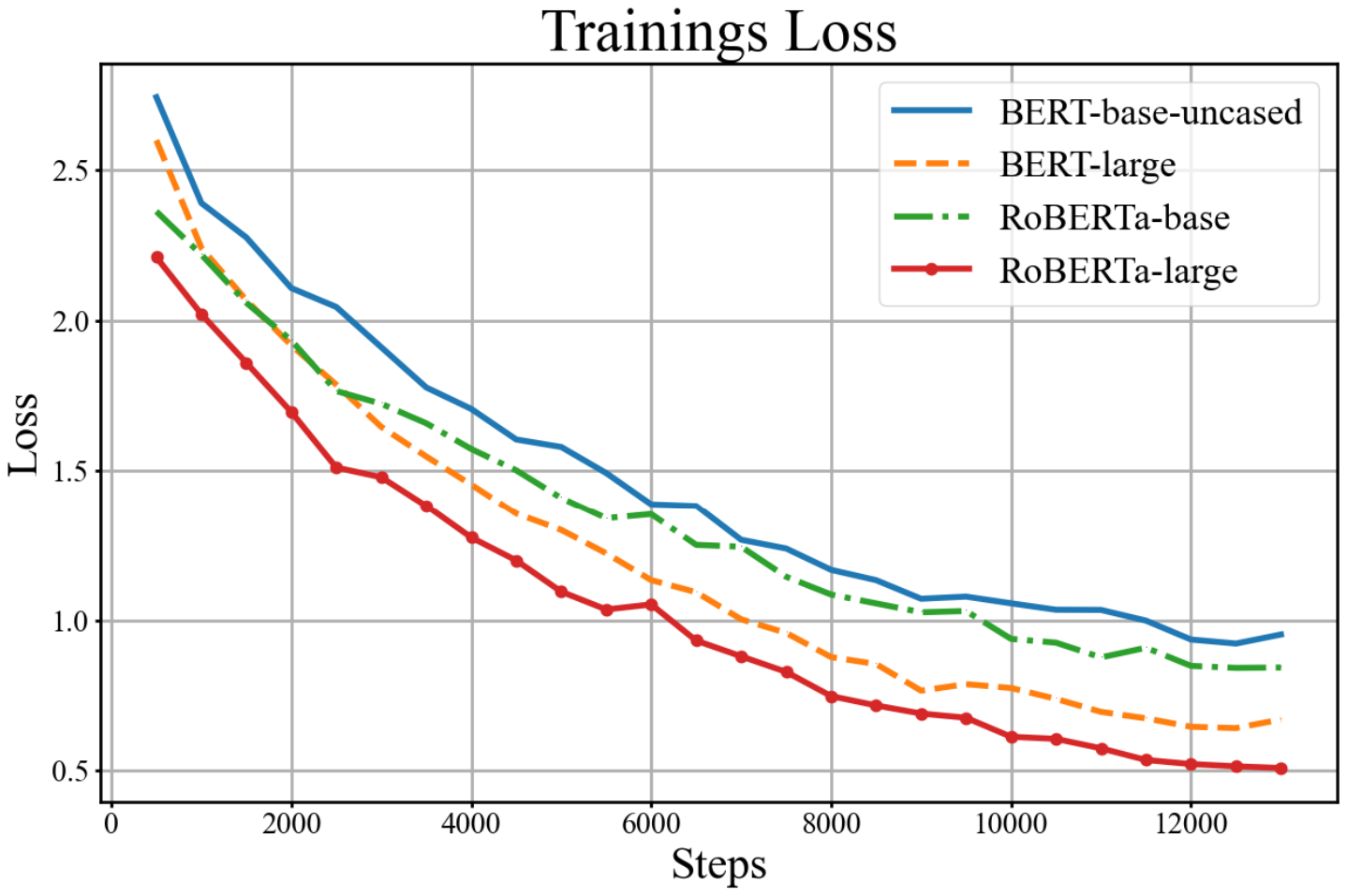}}
  \caption{\textbf{Training loss} of the four base models during continued pre-training on the original dataset.}
  \label{fig:Pretraining_trainings_loss}
\end{figure}

\begin{figure}[tb]
  \centering
  \makebox[\linewidth]{\hspace{-0.05\linewidth}\includegraphics[angle=-90,width=1.0\linewidth]{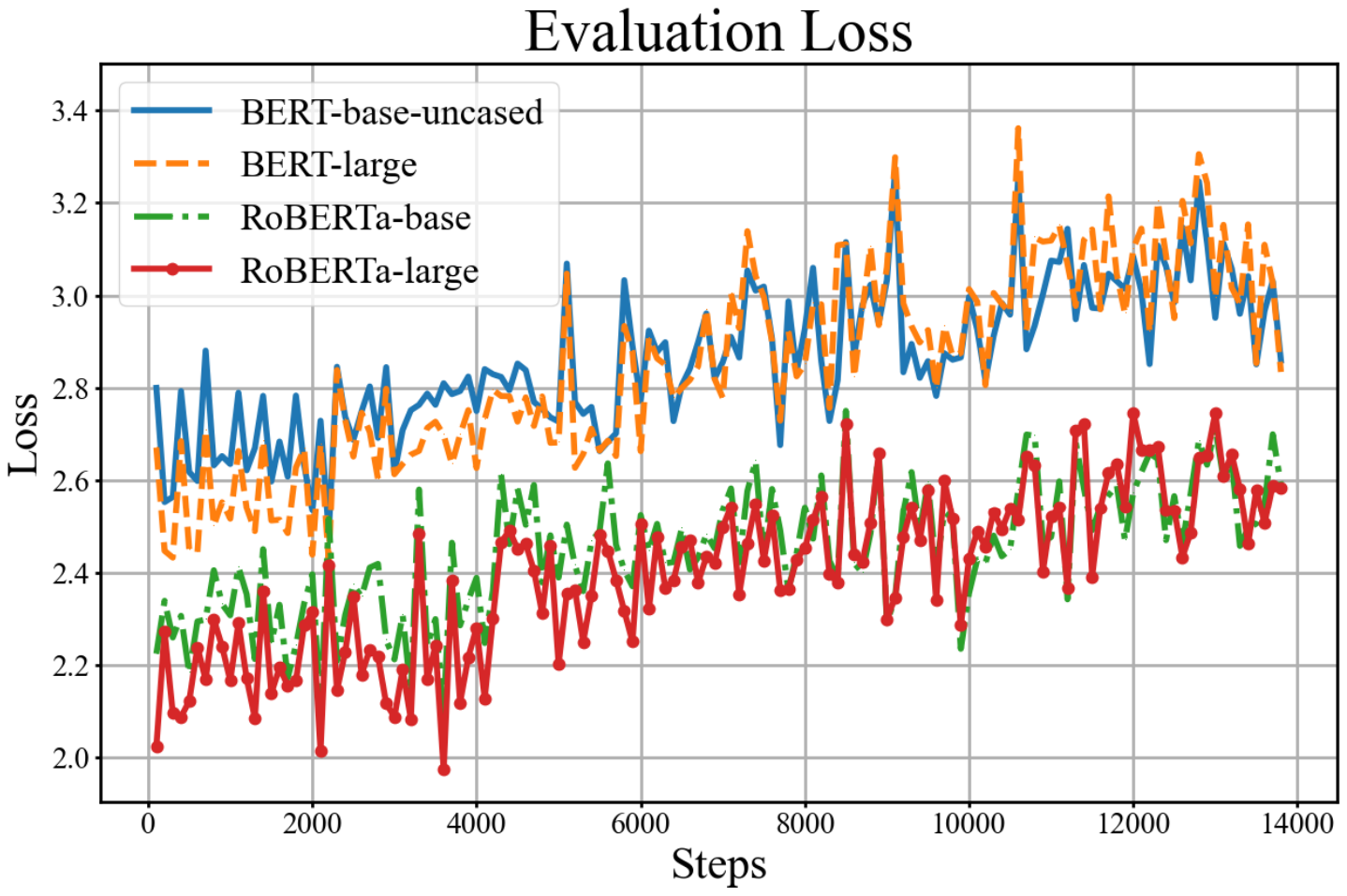}}
  \caption{Evaluation loss of the four base models during continued pre-training on the original dataset.}
  \label{fig:Pretraining_evaluation_loss}
\end{figure}

When comparing the evaluation loss (Figure \ref{fig:Pretraining_evaluation_loss}) with the training loss (Figure \ref{fig:Pretraining_trainings_loss}) during pre-training, it becomes evident that overfitting occurs rapidly, leading to diminishing returns. Specifically, continued domain-specific pre-training appeared to slightly improve the performance of the \code{bert-base} and \code{bert-large} models in most cases. However, the \code{roberta-base} and \code{roberta-large} models generally performed better without additional pre-training. This may be attributed to the fact that RoBERTa models were initially pre-trained for a longer duration on larger corpus, making continued pre-training prone to overfitting and a subsequent loss of generalization.

\paragraph{Impact of Generative Data Augmentation.}  

\begin{figure}[tb]
  \centering
  \makebox[\linewidth]{\hspace{-0.05\linewidth}\includegraphics[angle=-90,width=1.1\linewidth]{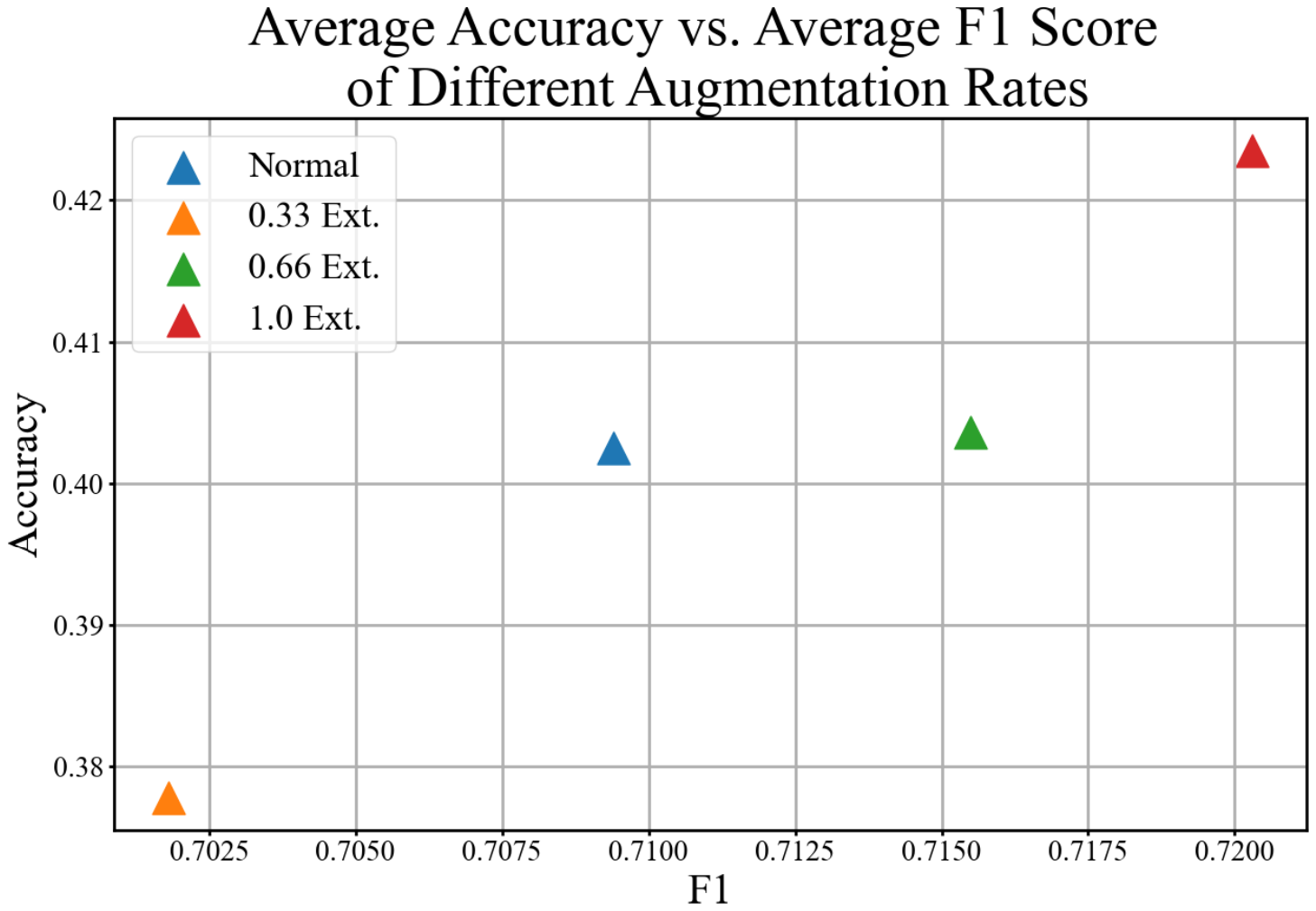}}
  \caption{Average F1 Scores and accuracy across different data augmentation rates.}
  \label{fig:Average Development}
\end{figure}

The evaluation results of the model for different data augmentation rates are shown in Figure~\ref{fig:Average Development} and Table~\ref{tab:model_performance}. Overall, accuracy and F1 scores increase as the training set expands, although gains are less than expected. Notably, smaller models such as \code{bert-base} show bigger performance improvements compared to larger models e.g., \code{roberta-large}.

A slight expansion of the training set initially results in a small drop in performance relative to the unmodified dataset. However, as the dataset size increases, performance improves: the dataset with 66\% augmentation performs comparably to the original, while the fully augmented dataset achieves the highest results—improving accuracy by two percentage points and F1 Score by an average of 0.01. In contrast, the \code{roberta-large} model consistently experiences performance degradation across all levels of data augmentation.

These findings suggest that the dataset size threshold, as described by \citeauthor{Balkus_Yan_2024}~\cite{Balkus_Yan_2024}, has not yet been reached. A further expansion of the training set might yield additional performance gains. Pre-training on the extended dataset appears to be less effective than direct fine-tuning, likely due to the introduction of noise into the domain-specific training data.

Overall, our results demonstrate that generative data augmentation is an effective strategy, which leads to a moderate improvement in model performance. By enriching the training domain with unmoderated examples generated by a state-of-the-art text generation model, we achieved measurable improvements in model performance. These findings underscore the potential of leveraging generative techniques to enhance the diversity and quality of training data.

\paragraph{Impact of Classification Head Architecture Variation.} 

The evaluation results of the model with different classification heads is summarized in Table~\ref{tab:roberta_large_heads_normal_vertical}. Overall, increasing the number of parameters in the classification head did not yield significant improvements in model performance. In fact, larger classification heads resulted in diminished performance, likely due to overfitting. Despite having fewer parameters, models utilizing the projected attention head demonstrated competitive effectiveness. This suggests that a well-designed attention-based classification head can enhance performance efficiency, achieving better results per parameter compared to standard fully connected layers.\\

\begin{table}
    \centering
        \begin{tabular}{lccc}
            \toprule
            \textbf{Class. Head} & \textbf{Accuracy} & \textbf{F1 Score} & \textbf{Params} \\
            \midrule
            fc 768x2            & \textbf{0.469}             & \textbf{0.77}             & 1x \\
            fc 768x4            & 0.417             & 0.743            & 3x \\
            fc 1536x2           & 0.458             & 0.77             & 4x \\
            fc 1536x4           & 0.455             & 0.756            & 12x \\
            Proj. Att. 128x1    & 0.432             & 0.743            & \textbf{0.44x} \\
            Proj. Att. 128x2    & 0.458             & 0.756            & 0.61x \\
            \bottomrule
        \end{tabular}
    \caption{Evaluation results of \code{roberta-large} model with different classification heads (fc--\textit{fully-connected} and Proj. Att--\textit{projected attention}).}
    \label{tab:roberta_large_heads_normal_vertical}
\end{table}

\subsection{Explainability Analysis Results}
\label{shap}

\begin{figure}[tb]
    \centering
    \includegraphics[width=0.75\linewidth]{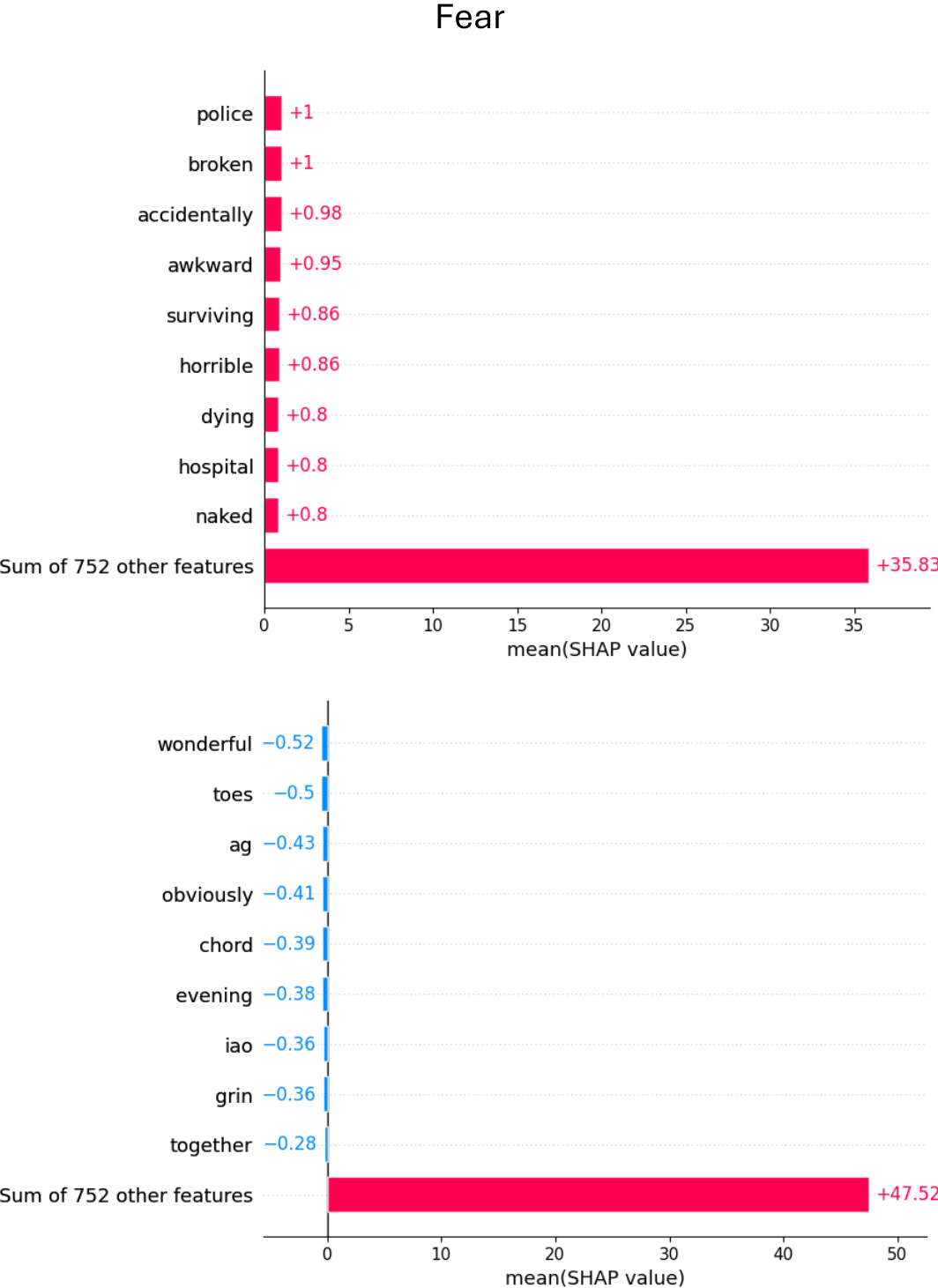}
    \caption{SHAP graph for ``Fear'' emotion class.}
    \label{fig:cropfear}
\end{figure}

\begin{figure*}[h]
    \centering
    \includegraphics[width=0.80\linewidth]{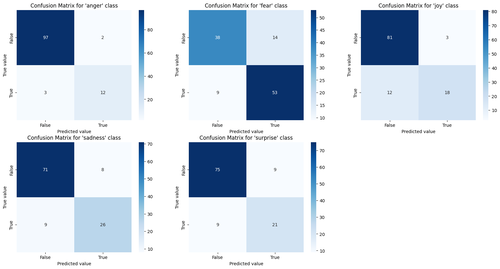}
         \vspace*{-0.2cm}
    \caption{Confusion matrix of the model.}
    \label{fig:confusion_matrixx}
\end{figure*}

Figure~\ref{fig:cropfear} shows the class-specific influence graph for the ``Fear'' class produced by SHAP analysis, with the graphs and explanation for other classes provided in the \textit{Appendix} 5.1. These visualizations offer insights into the key features influencing classification decisions by the model. For instance, the token \textit{police} emerges as a strong predictor for the ``Fear'' class, indicating that the model has learned to associate this token with expressions of fear in the dataset. These influence patterns serve as baseline indicators of the model’s interpretability. The SHAP analysis further reveals that the models construct a structured representation of sentiment, leveraging specific words and word combinations to make predictions.

Furthermore, the confusion matrices in Figure~\ref{fig:confusion_matrixx} offer deeper insights into the challenges of sentiment classification in short texts. They reveal that classes with higher prior probabilities are more accurately classified, while low-frequency classes suffer from reduced accuracy, reflecting the model’s reliance on data distribution in making the decision. For instance, the high number of false positives for the ``Fear'' class suggests a learned bias, providing insight into how the model internalizes and applies decision boundaries. This underscore how both training data quality and the complexity of emotional expression influence model interpretability.

\subsection{Reviewer Agreement and Human Evaluation}

Although an F1 Score below 0.9 and an accuracy under 80\% may initially seem suboptimal, it is crucial to recognize that multi-class sentiment analysis is inherently subjective. The high-dimensional nature of sentiment data and the limited long-context understanding in short texts often prevent exceptionally high accuracy. To further investigate this subjectivity, we conducted a human evaluation in which three English speakers (\textit{non-native}) annotated a subset of texts from the development set and compared it against the gold truth. This assessment provided insights into annotation variability and the inherent challenges of sentiment classification.

The results, as summarized in Table~\ref{tab:human_model_comparison1} and Table~\ref{tab:human_model_comparison2}, show that our model significantly outperforms human annotators, demonstrating its robustness in handling nuanced sentiment distinctions. Notably, our best model, despite an accuracy of only 0.469 and an F1 Score of 0.77, outperformed human evaluators, underscoring its capability in sentiment classification. Moreover, the moderate Cohen’s Kappa Scores (ranging from 0.293 for "Fear" to 0.709 for "Joy") highlight the difficulty of achieving consistent sentiment annotations. Lower agreement for emotions such as "Fear", \textit{Sadness}, and \textit{Surprise} suggests that these categories are inherently more ambiguous, particularly for non-native speakers. This variability further supports our observation that the model effectively navigates the complexities and subjectivity inherent in sentiment classification. 

\begin{table}[tb]
    \centering
    \begin{tabular}{lcc}
        \toprule
         & \textbf{Accuracy} & \textbf{F1 Score} \\
        \midrule
        Human Evaluator 1 & 0.224 & 0.549 \\
        Human Evaluator 2 & 0.29 & 0.612 \\
        Human Evaluator 3 & 0.252 & 0.578 \\
        Our Model         & 0.469 & 0.77 \\
        \bottomrule
    \end{tabular}
    \caption{Human evaluations vs. our model.}
    \label{tab:human_model_comparison1}
\end{table}

\begin{table}[tb]
    \centering
    \begin{tabular}{lcc}
        \toprule
         & \textbf{Cohen's Kappa} \\
        \midrule
        Anger & 0.504\\
        Fear & 0.293 \\
        Joy & 0.709 \\
        Sadness & 0.307 \\
        Surprise & 0.346 \\
        \bottomrule
    \end{tabular}
    \caption{Cohen's Kappa Score between human evaluators.}
    \label{tab:human_model_comparison2}
\end{table}

\section{Conclusion}

In this study, we investigated the impact of three key factors -- (1)~continued domain-specific pre-training, (2) generative data augmentation (GDA), and (3) classification head architecture—on multi-label sentiment classification in short texts. Given the limited contextual information typical of short-text datasets, these factors are especially critical for improving the performance of \textit{small} Transformer-based language models e.g., BERT and RoBERTa.

Our results indicate that while larger-parameter models (e.g., \code{roberta-large}) generally outperform smaller-parameter models (\code{bert-base} and \code{roberta-base}), moderate application of GDA consistently enhances the performance of smaller models, suggesting that synthetic samples effectively mitigate data scarcity. continued pre-training was beneficial for BERT-based architectures but introduced noise in RoBERTa variants, highlighting the model-specific nature of domain adaptation for short texts. Although modifications to the classification head had minimal impact on accuracy, they may reduce trainable parameters -- an important consideration for resource-constrained setting common in real-world applications. 

Our findings advance the development of tailored learning methods to overcome practical data difficulties, offering actionable insights for optimizing multi-label sentiment classification in short texts—a domain characterized by sparse and ambiguous data. Future work could explore refined filtering strategies for GDA and evaluate a broader range of architectures to further improve performance, particularly in resource-limited and complex data scenarios.

\bibliography{references}

\subsection{Appendix}
The following shows the class-specific influence graph for each sentiment class produced by SHAP analysis. As explained in \S\ref{shap}, these visualizations reveal the key features that influence the predictions of the model for each class, as described below. 

For the “Anger” class, \textit{insulting} and \textit{furious} are a strong positive predictor, while \textit{ailments} and \textit{laughed} contribute minimally. In the “Joy” class, \textit{wonderful} and \textit{thank} are a strong positive predictor, whereas \textit{horrible} and \textit{racism} are a strong negative predictor. 
For the “Surprise” class, \textcolor{black}{\textit{gt}\footnote{Here, \textit{gt} is an artifact from scraped data (representing the “greater than” symbol in HTML encoding), likely introduced during web scraping. Its presence is spurious and should be disregarded in the interpretation.}
followed by \textit{reality}, \textit{something}, and \textit{accidental} appear as a strong predictor,} while \textit{tears} and \textit{heavy} have minimal influence.
These findings indicate that the model has learned to associate these tokens with class-specific expressions present in the dataset. The influence patterns from the SHAP analysis serve as baseline indicators of the model’s interpretability.

\begin{figure}[h]
    \centering
    \includegraphics[width=0.7\linewidth]{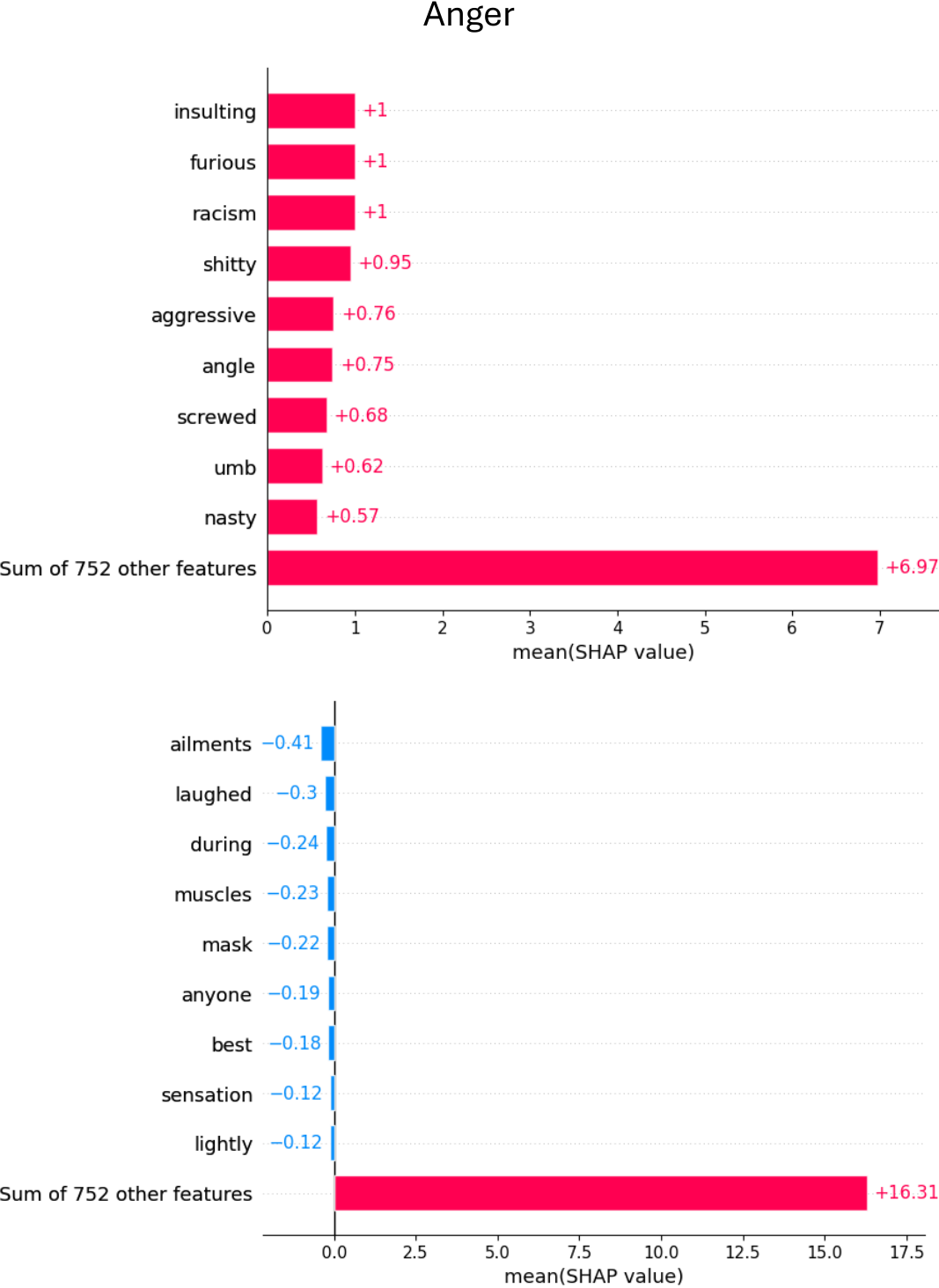}
    \caption{SHAP graph for ``Anger'' emotion class}
\end{figure}

\begin{figure}[h]
    \centering
    \includegraphics[width=0.7\linewidth]{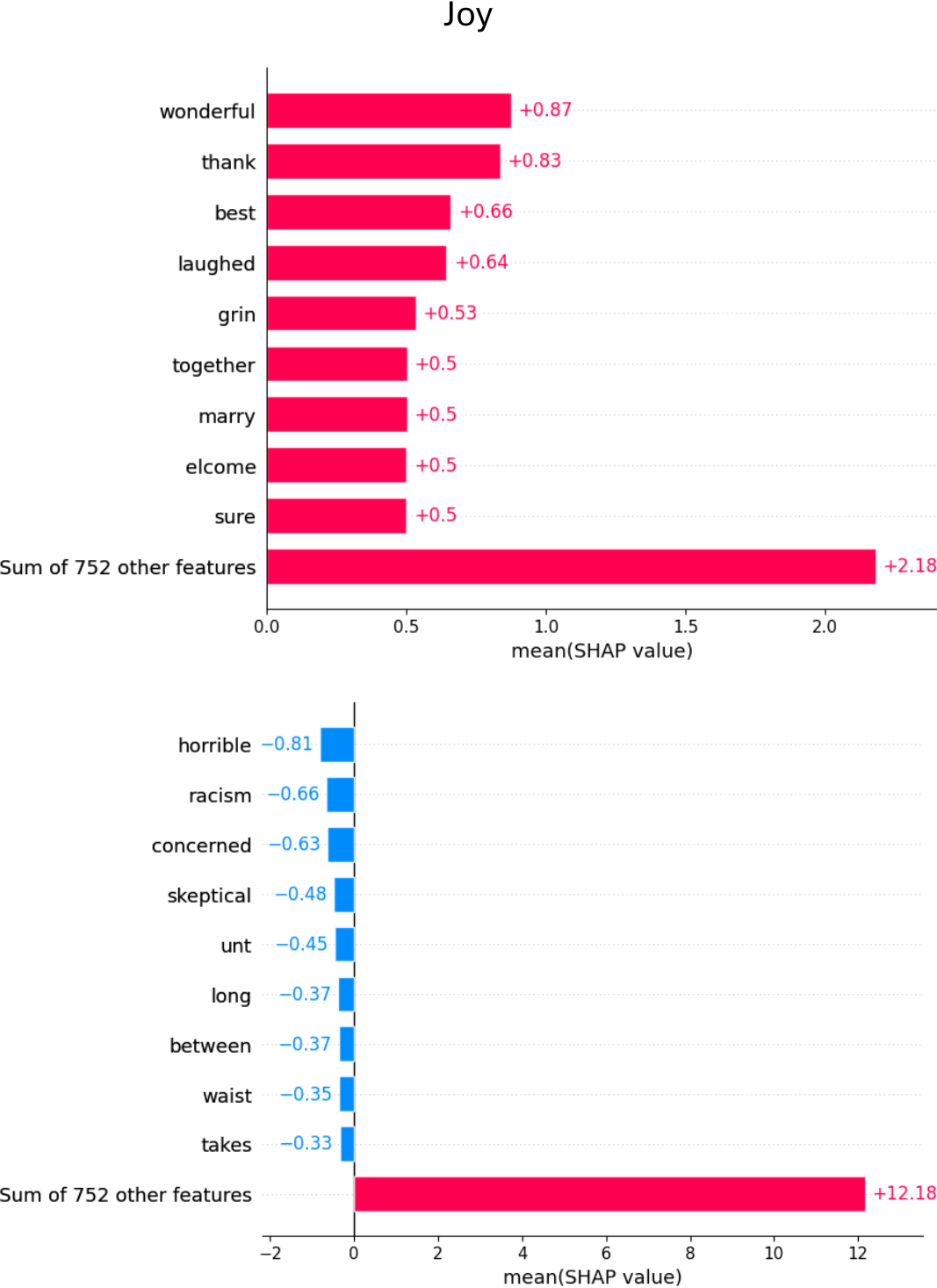}
    \caption{SHAP graph for ``Joy'' emotion class}
\end{figure}

\begin{figure}[h]
    \centering
    \includegraphics[width=0.7\linewidth]{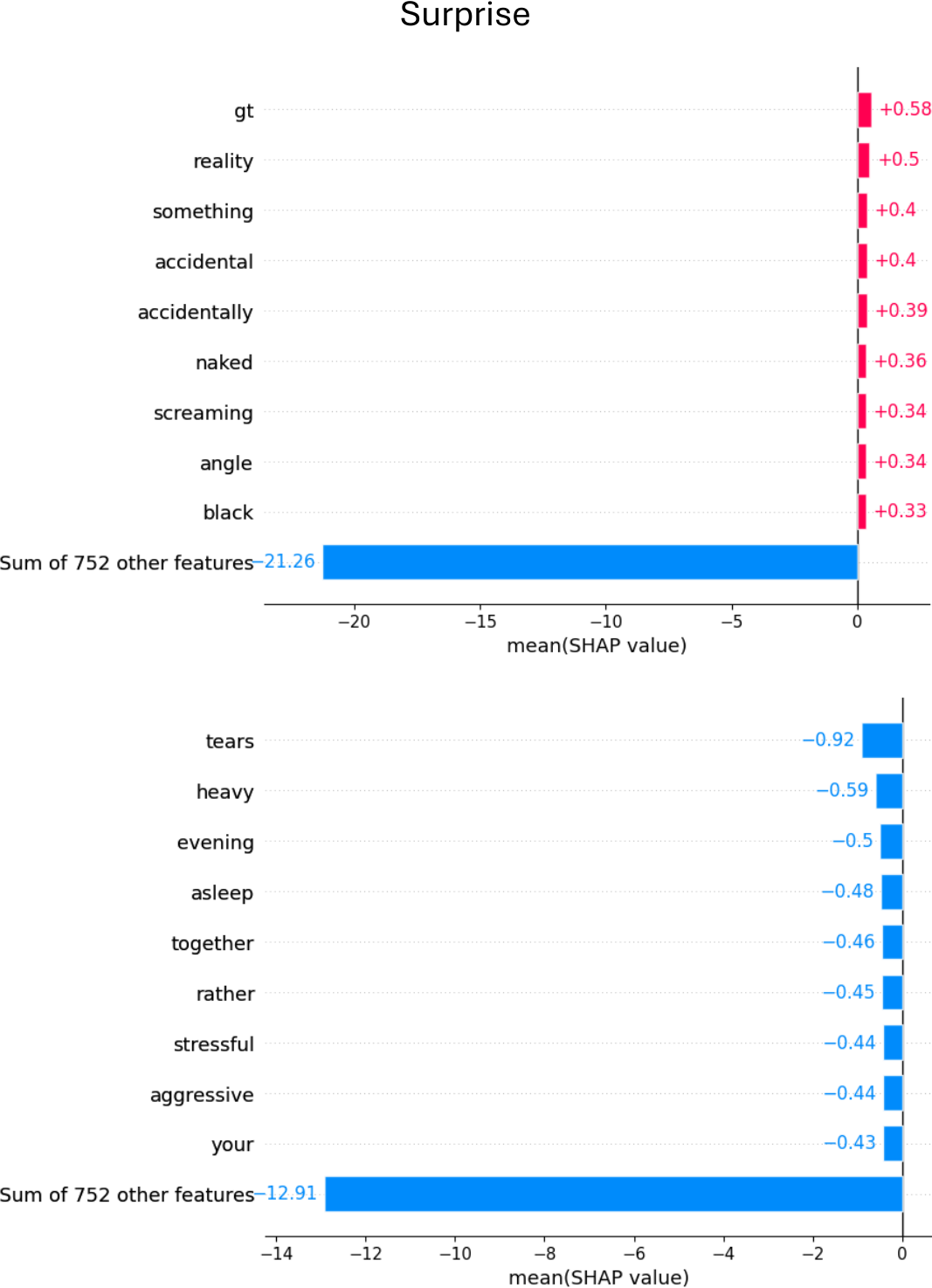}
    \caption{SHAP graph for ``Surprise'' emotion class}
\end{figure}

\end{document}